%% file: main.tex
\documentclass[letterpaper]{article}
\usepackage[]{aaai25}
\usepackage{times}
\usepackage{helvet}
\usepackage{courier}
\usepackage[hyphens]{url}
\usepackage{graphicx}
\graphicspath{{figures/}}
\usepackage{booktabs}
\usepackage{multirow}
\usepackage{array}
\usepackage{amsmath}
\usepackage{microtype}
\usepackage{natbib}
\usepackage{caption}
\DeclareCaptionStyle{ruled}{labelfont=normalfont,labelsep=colon,strut=off}
\usepackage{amssymb}
\usepackage{xcolor}
\input{macros/tpp_public_macros.tex}
\input{macros/case_table_macros.tex}

\title{Cross-Platform Production Signatures of State-Linked Information Operations}

\author{Lin Tian \& Marian-Andrei Rizoiu \\
  Behavioral Data Science, University of Technology Sydney \\
  \texttt{\{Lin.Tian-3,Marian-Andrei.Rizoiu\}@uts.edu.au}
}
\affiliations{}
\nocopyright %

\begin{document}
\maketitle

\input{sections/00_abstract}
\input{sections/01_introduction}

\input{sections/02_related_work}
\input{sections/03_methodology}

\input{sections/04_data_and_design}
\input{sections/05_rq1_transfer}
\input{sections/06_rq2_specificity}

\input{sections/07_rq3_context}
\input{sections/08_telegram}
\input{sections/09_conclusion}

\bibliography{references}

\appendix
\input{sections/appendix}

\end{document}

%% file: macros/tpp_public_macros.tex
\newcommand{\publicTPPPearson}{0.84}
\newcommand{\publicTPPSpearman}{0.81}

%% file: macros/case_table_macros.tex
\newcommand{\publicAdaptationRows}{%
ru-IRA & 1,323 & 30 & 0.84 & 1.00 \\
en-IRA & 1,480 & 1 & 0.85 & 0.99 \\
BLM & 28 & 4.5 & 0.94 & 0.83 \\
MAGA & 35 & 1 & 0.91 & 0.91 \\
}

%% file: sections/00_abstract.tex
\begin{abstract}
An information operation active on several platforms presents different
names, personas, and topics on each. Its working routine is harder to
change. We call this routine the operation's \emph{production signature}:
the regularities in when and how its accounts post, reply, and repost. We
measure it from timestamps and coarse action types only, using no content,
profile metadata, or network ties, as timing and action profiles for
accounts and learned action--gap policies for populations. We then ask which
parts of a signature carry over to another platform and which belong to a
single operation. We train models on Twitter archives of four state-linked
operations and apply them to Internet Research Agency (IRA) accounts on
Reddit, with no Reddit training labels. They rank the IRA accounts at AUC
0.90--0.94. Content detectors on the same route fall from 0.98 within a
platform to 0.81. Daily phase accounts for most of the transfer. Posting
volume reverses direction between platforms, and action mix follows the role
accounts hold on each platform. What transfers is not unique to the IRA: a
second operation shares the same working-hours window, and action--gap
policies separate every operation from organic accounts (AUC 0.96--1.0)
without consistently ranking their own operation first. On 938 Telegram
channels, groups that forward the same sources keep ordinary hours and pace
but publish original posts in the same ten-minute windows more often than
chance. This pattern is invisible when accounts are examined one at a time.
Our results suggest that timing and action traces learned on one platform
can help flag likely operations on another.
\end{abstract}

%% file: sections/01_introduction.tex
\section{Introduction}
\label{sec:intro}

On May 21, 2016, protesters and counterprotesters confronted each other
outside an Islamic center in Houston. One side had answered a Facebook call
from ``Heart of Texas,'' the other from ``United Muslims of America.'' The
shouting was real; neither organizer existed. Both pages were run by Russia's
Internet Research Agency (IRA) from the same St.~Petersburg operation, an
ocean away from the street they had filled
\citep{allbright2017houston,diresta2018,howard2018}.\footnote{The
advertisements for both events are in the House Intelligence Committee's
archive of IRA-purchased Facebook advertisements,
\url{https://democrats-intelligence.house.gov/facebook-ads/social-media-advertisements.htm}.}
The same agency was at work on Reddit in the same years, under a different
kind of persona. 

In April 2018 Reddit listed 944 accounts it suspected of
IRA ties. They had made more than 14,000 posts, most often not in political
communities but in r/funny, and the most successful of them, opened in
2015, had collected about 99,000 karma from jokes, memes, and partisan
posts across r/funny, r/PoliticalHumor, and
r/The\_Donald.\footnote{Reddit's announcement and 2017 transparency
report: \url{https://www.reddit.com/r/announcements/comments/8bb85p/} and
\url{https://redditinc.com/policies/transparency-report}; account-level
reporting:
\url{https://techcrunch.com/2018/04/11/reddit-bans-944-accounts-linked-to-russian-internet-research-agency}
and
\url{https://www.newsweek.com/reddit-account-has-karma-score-99493-and-it-has-just-been-suspended-amid-880106}.}
A protest page in Texas and a joke account on Reddit
differed in cause, audience, persona, and platform, yet shared an
organizational source. These cases illustrate state-linked information
operations (IOs): coordinated efforts by government-backed actors to shape
public opinion through inauthentic accounts
\citep{woolley2018,starbird2019,badawy2018,zannettou2019}. What remains
recognizable when an operation changes its language, topic, persona, or
platform?

\begin{figure}[t]
\centering
\includegraphics[width=\columnwidth]{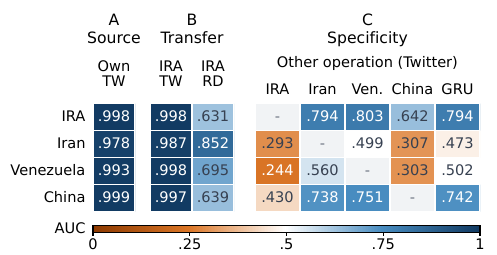}
\caption{Action--gap scores learned from each row's Twitter operation.
Account-level AUC compares (A) held-out source versus organic accounts,
(B) IRA versus organic accounts on Twitter (TW) and Reddit (RD), and
(C) source versus other Twitter operations. White marks 0.5; orange in C
means the other operation tends to score higher.
Protocol and 95\% intervals: Appendix Figure~\ref{fig:rep-task}.}
\label{fig:transfer-gap}
\end{figure}

Existing approaches to IO detection draw on three main types of signals: account content and profile attributes \citep{addawood2019,im2020,alizadeh2020}; relational information, including interactions, shared URLs, and connections to known seed accounts \citep{pacheco2021coordinated,nizzoli2021coordinated,saeed2022}; and behavioral patterns in the types and timing of account actions \citep{luceri2020,nwala2023bloc,schneider2026beyondcontent}.
Prior work has evaluated content-based detection across Twitter and Reddit \citep{alizadeh2020} and linked activity across platforms through shared artifacts \citep{wilson2020,golovchenko2020,barbero2023}. 
Behavioral models, in contrast, have largely been evaluated on the
platform they were trained on. Whether a learned signal remains useful on
another platform, and which patterns are shared across operations, are
open questions. Figure~\ref{fig:transfer-gap} illustrates both with
action--gap scores fitted on four Twitter operations. Each score separates
its source operation from organic Twitter accounts (A), but its ranking of
IRA accounts against organic accounts changes on Reddit (B), and the same
score can rank a different operation above its own (C). A regularity can
mark activity as unlike the platform's reference without identifying the
operation that produced it, so portability across platforms and
specificity to an operation need to be tested separately.

An operation's words and personas are written for an audience and are
easily changed. Producing them leaves a second trace, in when its accounts
post, how soon they return, and which actions they take (expression
\emph{given off} rather than \emph{given}, in Goffman's terms;
\citealp{goffman1959}). Because operations are staffed workplaces with roles
and schedules \citep{starbird2019,linvill2020troll}, we expect this trace
to carry regularities that persist when the audience or platform changes.
We do not expect them to be unique to one operation, since ordinary
institutions keep regular schedules too \citep{zerubavel1981}. In this
paper, we model this trace as the operation's \emph{production signature}:
the regularities in when and how a population of accounts posts, replies,
and reposts, measured from timestamps and coarse action types alone (i.e.,
without content, profile metadata, or network ties) on one platform over a
stated period. We then extend the analysis to
Telegram, whose forwarding records make relations between channels
observable, and ask whether channels that share forwarding sources also
publish original posts together. 

We focus on three research
questions:
\textbf{RQ1:} \emph{Which components of a production signature transfer
across platforms?} We test whether scores fitted on Twitter rank IRA
accounts above organic accounts on Reddit, with no Reddit labels used for
training, and which components carry the transfer.
\textbf{RQ2:} \emph{How specific are signatures to an operation?} We test
whether a score fitted on one operation ranks that operation above the
others, and which timing and action patterns operations share.
\textbf{RQ3:} \emph{How do signatures change within an operation?} We
measure which components change over time and across audiences, and
whether resemblance between populations extends to linked individual
accounts.

%% file: sections/02_related_work.tex
\section{Related work}
\label{sec:related}

\begin{figure*}[t]
\centering
\includegraphics[width=\linewidth]{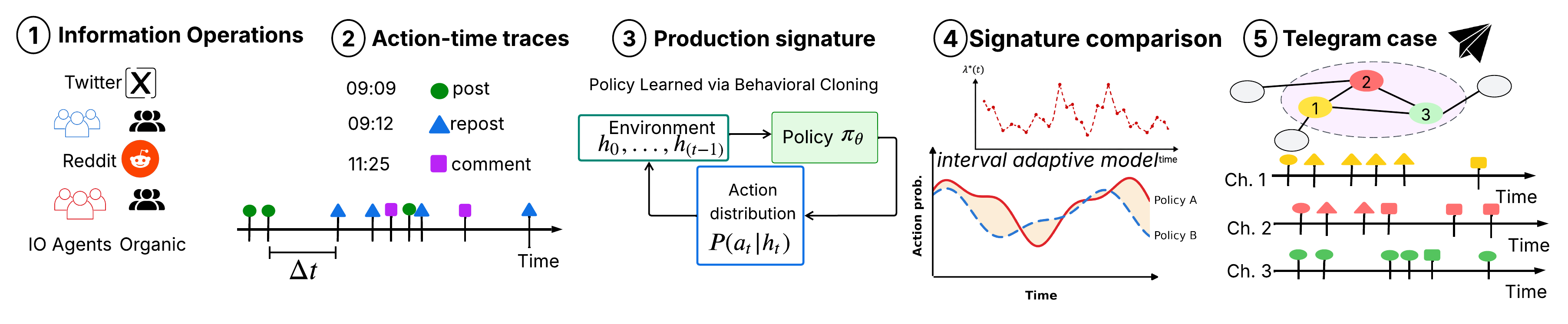}
\caption{Production signature framework for comparing the organizational
behavior of information operations across platforms.}
\label{fig:pipeline}
\end{figure*}

\paragraph{Computational detection of information operations.}
Attributions of state-linked operations come from platform disclosures and
investigative reporting, which cover only the accounts a platform releases or
a reporter finds \citep{gadde2018,diresta2018}. Computational methods aim to
find operation accounts from observable activity at scale, and they differ
mainly in what they observe. Supervised classifiers identify operation accounts and posts using linguistic
features, account metadata, and shared links
\citep{addawood2019,im2020,alizadeh2020}. Network methods instead group
accounts by shared artifacts and coordinated activity
\citep{pacheco2021coordinated,nizzoli2021coordinated}, while TrollMagnifier
uses interactions with known trolls to detect additional accounts on Reddit
\citep{saeed2022}. Behavioral approaches learn from accounts' actions:
\citet{luceri2020} infer policies through inverse reinforcement learning,
and \citet{ezzeddine2023} use an LSTM to classify sequences of actions and
received feedback. Related bot-detection methods exploit correlated activity
\citep{chavoshi2016} or cluster temporal representations learned by an LSTM
autoencoder \citep{mazza2019rtbust}. BLOC encodes actions, pauses, and content signals as symbolic sequences \citep{nwala2023bloc}. On Reddit, learned policies
have been used to measure behavioral homophily \citep{yuan2025behavioralhomophily}
and distinguish information operation accounts \citep{schneider2026beyondcontent}.
These studies establish several ways to represent behavior, but leave open
whether timestamps and coarse actions support operation detection across social platforms.

\paragraph{Cross-platform transfer and linkage.}
\citet{alizadeh2020} evaluate content-based information operation detection
across periods, countries, and platforms, including transfer between Twitter
and Reddit. Behavioral detection has also been tested across campaigns within
Twitter \citep{ezzeddine2023}. Work on bot classifiers shows that generalization
to unseen bot populations depends on the training data and evaluation design
\citep{echeverria2018lobo,yang2020scalable}. Other cross-platform studies trace
campaigns through shared narratives and videos
\citep{wilson2020,golovchenko2020}, multimodal embeddings
\citep{barbero2023}, and coordination networks \citep{cinus2025}.
These studies connect activity observed on multiple platforms; they do not
by themselves establish whether an account's behavior remains discriminative
under a platform shift. We fill in this gap by learning event-specific behavioral policies through behavioral cloning.

\paragraph{Temporal rhythms and organized production.}
Research on social time describes how schedules organize everyday activity
\citep{zerubavel1981}, while studies of online expression and human activity
document daily cycles and bursty timing \citep{golder2011,goh2008}.
Studies of information operations describe collaborative work, audience-specific
personas, and specialized roles \citep{starbird2019,arif2018,linvill2020troll,king2017}.
Coordination patterns can also reveal relations between producers
\citep{schoch2022}. These studies motivate us to find out which
behavioral regularities persist when the setting changes.

%% file: sections/03_methodology.tex
\section{Methodology}
\label{sec:method}

Following Figure~\ref{fig:pipeline}, we construct account action--time traces,
learn population-level behavioral policies, and adapt temporal comparisons to
each platform through interval-adaptive modeling. The adaptation stage fits
a reference on each platform and rescales observed gaps against that reference.
We then use policy scores, relative-tempo scores, and distances to compare
accounts and populations. Component profiles summarize phase, schedule, tempo,
and action mix so that we can examine which regularities carry these comparisons.

\paragraph{Action--time traces.}
Let $p$ denote a platform, $u$ an account on that platform, and $n_u$
its number of observed events in the analysis window. Its trace is
$X_u^{(p)}=((t_{ui},z_{ui}))_{i=1}^{n_u}$, where timestamps $t_{ui}$
are ordered UTC Unix seconds and $z_{ui}=g_p(a_{ui})$ maps native action
$a_{ui}$ to $\mathcal A=\{\mathrm{post},\mathrm{reply},\mathrm{repost}\}$.
For $i\ge2$, $\Delta t_{ui}=(t_{ui}-t_{u,i-1})/60$ is the gap in
minutes. Only the interval-adaptive model collapses exact timestamp ties.
Nothing else about an account enters a behavioral score: content, ties,
handles, and metadata serve only to select cohorts. Throughout, $P$ and
$Q$ denote populations on a stated platform and period.

\subsection{Production-signature profiles and policies}

\paragraph{Component profiles.}
\label{sec:method-features}
We model each account's profile with four feature groups. \emph{Phase} comprises
24 normalized UTC-hour counts $h_u$, with nonnegative entries summing to
one, and three fixed UTC-window masses. \emph{Schedule} comprises the
weekday/weekend activity ratio, the sum of the six largest hourly proportions,
hour entropy, and active-day density. The six hours need not be contiguous.
\emph{Tempo} comprises median and 90th-percentile gaps, burstiness
\citep{goh2008}, and the fraction of gaps below one minute, including ties.
\emph{Action mix} comprises the three observed action proportions. The
full profile combines all four groups, the phase-free profile
omits phase (it is not fully timezone-invariant, because shifts can change
weekday assignments and hourly concentration), and the earlier
phase + weekday specification uses phase and the weekday ratio. 
Feature definitions are in the Appendix.

\paragraph{Action--gap policies.}
\label{sec:method-policy}
We model each population's action--gap policy as a next-token
distribution over action--gap tokens, where each token combines an action
with the category of the gap that preceded it:
\begin{equation*}
c_{ui}=\big(z_{ui},q(\Delta t_{ui})\big),\qquad i=2,\ldots,n_u.
\end{equation*}
The quantizer $q$ uses empirical gap-decile cutpoints, with duplicate
cutpoints merged; the vocabulary allocates $3\times10=30$ tokens.
The first event supplies context $c_{u1}=(z_{u1},q(0))$, whose zero gap is
an initialization convention. Each population's behavior is characterized
by a policy describing its distribution over the next token given the
preceding ones. Formally, the policy of population $P$ is
$p_{\widehat\psi_P}(c_{ui}\mid H^c_{ui})$, the distribution of the next
token $c_{ui}$ given the history $H^c_{ui}$ of tokens seen since the last
state reset, with parameters $\widehat\psi_P$ fitted by behavioral
cloning, that is, next-token prediction on the traces of $P$'s accounts,
using a two-layer Mamba \citep{gu2024mamba}. The policy represents the
population's learned production pattern.

\subsection{Interval-adaptive modeling}
\label{sec:method-tpp}
We model tempo relative to each platform with an \emph{interval-adaptive
model}: it scores each gap over an interval that accommodates timestamp
precision, and it adapts to each platform by learning a reference there.
First, a
Mamba encoder \citep{gu2024mamba} learns a conditional gap distribution
from binned gap histories on each platform's reference population; it is
trained on the probability mass of each gap's scoring interval, an
interval-probability objective inspired by interval-censored Hawkes
estimation \citep{hawkes1971,rizoiu2022ichawkes}, with a mixture
cumulative distribution function (CDF) in place of an explicit excitation
kernel. Second, the fitted reference maps each gap's interval to a
randomized residual through its conditional CDF
\citep{brown2002timerescaling}, so that the same gap in minutes can
receive a different relative tempo on different platforms. Third, a
model fits each population's residual sequences. 

\paragraph{Interval-probability objective.}
Let $\widetilde t_{u1}<\cdots<\widetilde t_{um_u}$ be account $u$'s
$m_u$ distinct timestamps after collapsing ties. Index its gaps by
$j=1,\ldots,m_u-1$ and define
\begin{equation*}
\begin{aligned}
d_{uj}&=\left\lfloor
  \frac{\widetilde t_{u,j+1}-\widetilde t_{uj}}{60}\right\rfloor,\\
L_{uj}&=\max\{d_{uj}-1,\epsilon_t\},\qquad R_{uj}=d_{uj}+1.
\end{aligned}
\end{equation*}
Here $d_{uj}$ is the completed-minute gap, $(L_{uj},R_{uj}]$ its
scoring interval in minutes, and $\epsilon_t=10^{-3}$ minutes a numerical
floor. Distinct sub-minute events remain as $d_{uj}=0$. 

A Mamba-based encoder with parameters $\eta$ outputs a mixture of $K=8$
log-normal CDFs, denoted $F_\eta(x\mid H^d_{uj})$ for a positive gap
$x$ in minutes. The history $H^d_{uj}$ contains only preceding gaps,
encoded in 32 logarithmic bins, with encoder state reset at segment boundaries.
For retained training gaps $\mathcal I_{\mathrm{tr}}$, whose count is
$M=|\mathcal I_{\mathrm{tr}}|>0$, the interval mass and loss are
\begin{equation*}
\begin{aligned}
w_{\eta,uj}
 &=F_\eta(R_{uj}\mid H^d_{uj})-F_\eta(L_{uj}\mid H^d_{uj}),\\
\mathcal L_{\mathrm{int}}(\eta)
 &=-\frac{1}{M}\sum_{(u,j)\in\mathcal I_{\mathrm{tr}}}
       \log\max\{w_{\eta,uj},10^{-9}\}.
\end{aligned}
\end{equation*}

\paragraph{Automatic platform adaptation.}
Fit $F_p^0$ on training accounts from platform $p$'s reference
population: organic Twitter, organic Reddit, or the Telegram broadcast
reference. 
Let $A_{uj},B_{uj}$ be its CDF values at $L_{uj},R_{uj}$,
conditional on $H^d_{uj}$, after the numerical safeguards in the appendix.
One archived seeded sequence of
independent draws $V_{uj}\sim\mathrm{Uniform}(0,1)$ gives
\begin{equation*}
\begin{aligned}
U_{uj}&=A_{uj}+V_{uj}(B_{uj}-A_{uj}),\\
\widetilde U_{uj}&=\operatorname{clip}(U_{uj},10^{-9},1-10^{-7}),\\
\tau_{uj}&=-\log(1-\widetilde U_{uj}).
\end{aligned}
\label{eq:platform-rescaling}
\end{equation*}
Randomization across CDF jumps gives a uniform probability integral transform
(PIT) when the jumps are the true conditional probabilities of the observed
categories \citep{dunn1996randomized}. The negative-log-survival transform
then gives unit-exponential residuals \citep{brown2002timerescaling}.
Uniform and unit-exponential distributions are comparison references;
held-out Kolmogorov--Smirnov (KS) distances
\citep{kolmogorov1933,massey1951} measure departure from them.

\paragraph{Relative-tempo density.}
For each population $Q$, a second two-layer, eight-component Mamba mixture
fits a density $f_Q$ to the relative-tempo residuals. Its history $H^\tau_{uj}$
contains preceding binned residual inputs, with state reset per scoring window. 
We use $\tau^*_{uj}=\max\{\tau_{uj},10^{-8}\}$ for density
fitting and scoring. $f_Q$ is applied to describe residual sequences relative to
$Q$'s fitted platform reference, rather than absolute gaps.

\subsection{Account scores and cohort comparisons}
\label{sec:method-readouts}
For an individual account, we
compute a policy score, a likelihood ratio that asks whether the account's
token sequence is predicted better by the operation's policy or by the
reference's, and a relative-tempo score, which asks the same of its
relative tempo. For populations, we compute the distance between
two fitted policies and compare cohort phase and tempo directly.

\paragraph{Policy score.}
Let $p_{\widehat\psi_P}$ and $p_{\widehat\psi_0}$ be the next-token
probabilities learned for population $P$ and its reference population.
For nonempty scored positions $\mathcal K_u$, use the same tokens and
history $H^c_{ui}$ under both models:
\begin{equation*}
r_P(u)=\frac{1}{|\mathcal K_u|}\sum_{i\in\mathcal K_u}
\log\frac{p_{\widehat\psi_P}(c_{ui}\mid H^c_{ui})}
{p_{\widehat\psi_0}(c_{ui}\mid H^c_{ui})}.
\end{equation*}
The history contains preceding tokens available after the current state reset,
not the full account history. For a complete scored trace,
$\mathcal K_u=\{2,\ldots,n_u\}$; a held-out segment uses only its
eligible prediction positions. Positive scores favor $P$'s predictions
over the reference. 

\paragraph{Distance between fitted policies.}
Let $E_Q(P)$ be held-out $P$'s negative log-likelihood (NLL) under
$Q$'s fitted model: average over predicted tokens per account or
account-era segment, then equally across accounts. Define symmetric excess NLL:
\begin{equation*}
\begin{aligned}
D(P,Q)&=\tfrac12\big[E_Q(P)-E_P(P)\big]\\
      &\quad+\tfrac12\big[E_P(Q)-E_Q(Q)\big].
\end{aligned}
\end{equation*}
Each bracket compares the other model with the population's own model on
identical observations and histories under a shared vocabulary. Units are
nats/token. 
Relative-tempo distances use the same formula with $f_P,f_Q$, pooling
retained held-out gaps; their units are nats/residual gap, with unequal
account weights.

\paragraph{Relative-tempo score.}
Let $\mathcal J_u$ contain all relative-tempo residuals in account $u$'s evaluated
record and $\mathcal I_u\subseteq\mathcal J_u$ the gaps retained in density
scoring windows. Accounts must have $|\mathcal I_u|>0$. The archived
interval-adaptive model's account score is
\begin{equation*}
\begin{aligned}
\ell_Q(u)
 &=\frac{1}{|\mathcal I_u|}\sum_{j\in\mathcal I_u}
       \log f_Q(\tau^*_{uj}\mid H^\tau_{uj})\\
 &\quad+\frac{1}{|\mathcal J_u|}\sum_{j\in\mathcal J_u}\tau^*_{uj}.
\end{aligned}
\label{eq:rescaled-density-score}
\end{equation*}
It subtracts mean unit-exponential log density, $-x$ at
$x>0$. 

\paragraph{Cohort phase and tempo.}
\label{sec:method-adapt}
For nonempty eligible cohorts $C,R$, average normalized hourly profiles
with equal account weight and compare their centroids:
\begin{equation*}
\bar h_C=\frac{1}{|C|}\sum_{u\in C}h_u,\qquad
\operatorname{cos}(C,R)=
\frac{\bar h_C^\top\bar h_R}{\|\bar h_C\|_2\|\bar h_R\|_2}.
\end{equation*}
Cohort tempo is the median of the accounts' median gaps, in minutes.

%% file: sections/04_data_and_design.tex
\section{Data and baselines}
\label{sec:data-experiments}

\paragraph{Datasets.}
\label{sec:data}
Twitter's election-integrity releases\footnote{\url{https://transparency.twitter.com/en/reports/information-operations.html}} provide accounts attributed to the IRA,
Iran, Venezuela, China, and GRU \citep{gadde2018}; a same-region diagnostic
adds the Russian Current Policy network \citep{twitter2020currentpolicy}.
For Reddit, we use public activity for 146 accounts from the platform's IRA
list.\footnote{\url{https://redditinc.com/policies/transparency-report}}
Table~\ref{tab:data} shows the dataset sizes and periods.
The Telegram set is an unlabeled 2024 sample of public channels from the
election dataset \citep{blas2025telegram}.

\begin{table}[t]
\centering\small
\setlength{\tabcolsep}{2.5pt}
\begin{tabular}{@{}>{\raggedright\arraybackslash}p{.88in}llrr@{}}
\toprule
Population & Platform & Span & \# Accounts & \# Actions \\
\midrule
IRA & Twitter & 2009--2018 & 3,479 & 8.77M \\
Iran & Twitter & 2010--2018 & 492 & 1.12M \\
Venezuela & Twitter & 2010--2018 & 744 & 8.95M \\
China & Twitter & 2007--2020 & 3,321 & 7.53M \\
GRU & Twitter & 2013--2020 & 30 & 26,440 \\
\midrule
IRA & Reddit & 2015--2018 & 146 & 16,240 \\
\midrule
\multirow{2}{*}{Organic} & Twitter & 2007--2020 & 2,186 & 255,740 \\
 & Reddit & 2015--2021 & 943 & 5.87M \\
\midrule
Public channels & Telegram & 2024 & 938 & 5.86M \\
\bottomrule
\end{tabular}
\caption{Dataset statistics.}
\label{tab:data}
\end{table}

\paragraph{Baselines.}
\label{sec:baselines}
Table~\ref{tab:main-results} compares production signatures with content-
and behavior-based baselines. Baselines we fit on Twitter use the same
training accounts as our models and are evaluated on Reddit without
adaptation; within-Reddit baselines use the shared five-fold splits.
Content baselines include the reported content-and-profile logistic
regression of \citet{im2020}, logistic regression on multilingual MPNet
embeddings \citep{reimers2020multilingual} mean-pooled over each account's
posts, and XLM-R \citep{conneau2020xlmr} fine-tuned on posts with predictions
averaged per account. We also include the random-forest baseline of
\citet{alizadeh2020}.\footnote{The original paper reports only
macro-F$_1$. We train a 1{,}000-tree random forest on posts at a 1:2
troll-to-control ratio and average post probabilities per account.
Features comprise word, character, URL, hashtag, and mention counts;
20 LDA topics; fractions of top-25 troll and control words and bigrams;
URL-domain type, top-level domain, and top-25 troll and control domain
indicators; and posting hour and weekday.}
Behavior baselines include the reported classifier of \citet{luceri2020},
which uses rewards inferred from action sequences via inverse reinforcement
learning, and the empirical-policy
classifier of \citet{schneider2026beyondcontent}, reimplemented within
Reddit. BLOC
\citep{nwala2023bloc} encodes accounts as sequences of action symbols and
duration-binned pause symbols (under a minute, hour, day, week, month,
year, or longer), then applies logistic regression to bigram TF-IDF features.
All results are account-level AUC. 

%% file: sections/05_rq1_transfer.tex
\section{Results and Analysis}
\label{sec:results}
\label{sec:rq1}

\begin{table}[!t]
\centering\footnotesize
\renewcommand{\arraystretch}{0.92}
\begin{tabular*}{\columnwidth}{@{\extracolsep{\fill}}lr@{}}
\toprule
Model & AUC \\
\midrule
\multicolumn{2}{@{}l}{\textbf{TW$\to$TW: within-platform}} \\
\citet{luceri2020}$^{\dagger}$ & 0.891 \\
MPNet \citep{reimers2020multilingual} & 0.988 \\
XLM-R \citep{conneau2020xlmr} & 0.989 \\
\citet{im2020}$^{\dagger}$ & 0.989 \\
BLOC \citep{nwala2023bloc} & 0.998 \\
PS & \textbf{0.999} \\
Policy  & 0.998 \\
\midrule
\multicolumn{2}{@{}l}{\textbf{RD$\to$RD: within-platform}} \\
\citet{schneider2026beyondcontent}$^{\ddagger}$ & \textbf{0.993} \\
MPNet \citep{reimers2020multilingual} & 0.980 \\
BLOC \citep{nwala2023bloc} & 0.978 \\
PS  & 0.986 \\
Policy & 0.926 \\
\midrule
\multicolumn{2}{@{}l}{\textbf{TW$\to$RD: cross-platform transfer}} \\
\citet{alizadeh2020}$^{\ddagger}$ & 0.743 \\
XLM-R \citep{conneau2020xlmr} & 0.764 \\
MPNet \citep{reimers2020multilingual} & 0.812 \\
BLOC \citep{nwala2023bloc} & 0.918 \\
PS & \textbf{0.938} \\
\quad $-$ action mix & 0.863 \\
\quad $+$ phase, schedule & 0.907 \\
\quad $+$ phase & 0.924 \\
\quad $+$ schedule & 0.793 \\
\quad $+$ tempo & 0.718 \\
\quad $+$ volume & 0.665 \\
\quad $+$ action mix & 0.951 \\
\bottomrule
\end{tabular*}
\caption{Account-level AUC for detecting IRA accounts within a platform
(TW$\to$TW, RD$\to$RD) and across platforms with no target labels
(TW$\to$RD). PS $=$ production signature, each cell the better of
logistic regression and gradient boosting; indented rows drop ($-$) or use
only ($+$) the listed components. Policy $=$ learned policy via behavioral
cloning (BC). $^{\dagger}$Reported in the original paper.
$^{\ddagger}$Our reimplementation. Best result in each block in bold.}
\label{tab:main-results}
\end{table}

\begin{figure*}[t]
\centering
\includegraphics[width=\linewidth]{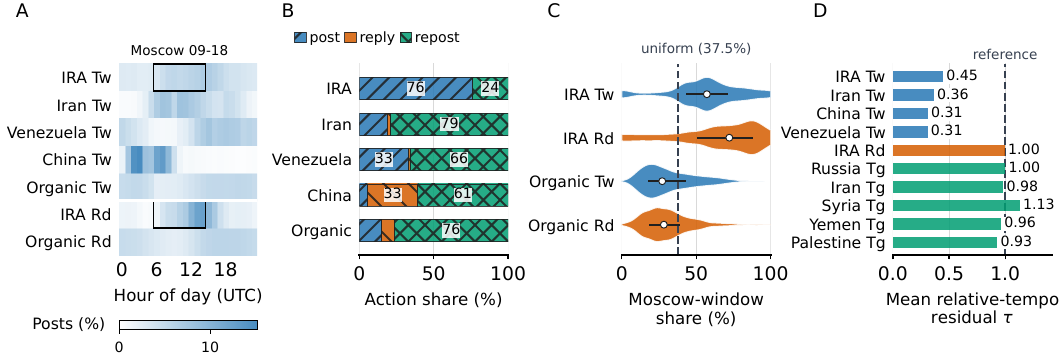}
\caption{Production signatures of four operations and two reference
populations. (A) Hour-of-day activity (UTC); boxes mark Moscow working
hours (09:00--18:00 UTC+3) on the IRA rows. (B) Median per-account action
shares on Twitter. (C) Per-account share of posts in Moscow working hours;
white points are medians and black bars the interquartile range.
(D) Mean relative-tempo residual per population, with the platform reference at
1 (dashed).}
\label{fig:clock-diagnostic}
\label{fig:ops}
\end{figure*}

\subsection{RQ1: Which components transfer across platforms?}
\label{sec:transfer}

Table~\ref{tab:main-results} reports account-level AUC for every model
fitted and tested within Twitter, within Reddit, and fitted on Twitter
then applied to Reddit without Reddit labels. The test cohort for the
transfer block is 96 Reddit IRA accounts against 168 volume-matched
controls from the 643-account reference cohort (the PS rows use 161
controls from a separate matching draw). All models perform well within each platform (0.89--1.00), so we focus on how well they transfer across platforms.

\paragraph{Transfer across platforms.}
The production signature transfers at AUC 0.938, the best cross-platform
result in the table. It gives up 0.05 relative to the same model fitted on
Reddit (0.986), and BLOC \citep{nwala2023bloc} gives up 0.06 (0.978 to
0.918). Content fares worse. MPNet drops 0.17 (0.980 to 0.812) and
fine-tuned XLM-R 0.23 (0.989 to 0.764). Our reimplementation of the
content-based random forest of \citet{alizadeh2020} is as accurate as the
other content models within a platform yet reaches only 0.743 across
platforms; at the post level its macro-F$_1$ is 0.55, close to the 0.60
they report for the same direction. The paired advantage of
phase + schedule over MPNet is $\Delta\mathrm{AUC}=+0.095$ with gradient
boosting and unresolved with logistic regression ($+0.046$). The accounts
change what they post to suit each audience and platform. Their posting
schedules stay much the same.

\paragraph{Transfer of phase.}
Phase alone ranks the IRA accounts at AUC 0.852/0.924 (logistic
regression/gradient boosting) and phase + schedule at 0.857/0.907, so the clock alone accounts
for most of the full profile's 0.938. IRA accounts concentrate in Moscow office
hours on Twitter and on Reddit alike, while the organic accounts around
them spread across the daily cycles that ordinary users keep
(Figure~\ref{fig:clock-diagnostic}A, C) \citep{golder2011}. The detector
finds scheduled production inside unscheduled activity, which fits
accounts of the IRA as a staffed workplace with specialized roles
\citep{linvill2020troll,starbird2019}; in Goffman's terms, the message is
expression given and the clock expression given off \citep{goffman1959}.
The result holds from the 2014 and 2016 source windows (AUC 0.79--0.95),
in the reverse direction from Reddit to Twitter (0.90), and from 2\% of
the source accounts (matched AUC 0.90), and sequence models given Reddit
labels do not improve on it (Appendix Table~\ref{tab:transfer}; Appendix
Figure~\ref{fig:detectors}).

\paragraph{Transfer of the other components.}
Tempo, the pace of posting, reaches only 0.694/0.718 and lowers
phase + schedule when added ($\Delta\mathrm{AUC}=-0.12$ to $-0.18$); the
advantage of phase + schedule over tempo ($+0.163$ and $+0.188$) keeps its
direction on the stricter gap-eligible cohort (component-ablation
appendix). Volume points the wrong way: IRA accounts out-post organic
accounts on Twitter and under-post them on Reddit, so a posts-per-day
score reverses to AUC 0.125, while the profile scores are nearly
uncorrelated with volume (Spearman 0.16 among IRA accounts, $-0.03$ among
organic accounts; Appendix Figure~\ref{fig:detectors}A). Action mix
transfers (matched AUC 0.951/0.927), but the ranking tracks the roles in
the target cohort: the Reddit IRA accounts are mostly submitters in a
comment-heavy reference, and action scores fitted on three operations
give the same 0.94 while the China score gives 0.06, because all four
order accounts by submission share. BLOC, whose symbols pair actions with
pauses of fixed duration, is the most stable baseline (0.918; 0.91 and
0.80 when fitted on Iran and China) but stays below the full profile and
below phase alone. How much an operation posts compared with other accounts, and the types
of posts it makes, vary across platforms. Its posting times remain stable,
consistent with routines theory \citep{feldman2003}.

\paragraph{Shared window and operating cutoff.}
If phase transfers because of the hours the IRA keeps, a phase score
fitted on any operation keeping similar hours should transfer as well.
Under a common protocol (representation-by-task appendix), the
Iranian-fitted phase score ranks Reddit IRA accounts above matched
controls at AUC 0.79/0.88, close to the IRA's own 0.86/0.91, while
China-fitted scores transfer weakly (0.61/0.74) and Venezuela-fitted
scores barely or in reverse (0.56/0.38) (Appendix
Figure~\ref{fig:rep-task}). The IRA and Iranian archives are the two whose
working-hours windows overlap (Figure~\ref{fig:ops}A). The learned
action--gap policies, near-perfect within Twitter and at 0.93 when
refitted within Reddit (Table~\ref{tab:main-results}), transfer just as
unevenly across source operations, from AUC 0.63 when fitted on the IRA to
0.85 when fitted on Iran (Appendix Figure~\ref{fig:rep-task}), which is
why the cross-platform block carries no policy row. The same shared
window limits the score in use. At the cutoff that flags 5\% of
Reddit reference accounts, the full profile catches 68\% of IRA accounts
with logistic regression and 43\% with gradient boosting, and
phase + schedule 49\% and 74\%; at 1\%, phase + schedule falls to 9\% and
47\%. The cutoff does not travel either: chosen on Reddit, it flags 99\%
of organic Twitter accounts. Flagged reference accounts keep
weekday-concentrated weeks (median weekday ratio 4.0 against 1.1 for
unflagged accounts), missed IRA accounts keep looser hours, and the
Moscow-window shares of the two populations overlap widely
(Figure~\ref{fig:clock-diagnostic}C). This may indicate institutional schedules organize
many populations at once \citep{zerubavel1981}.

%% file: sections/06_rq2_specificity.tex
\subsection{RQ2: How specific are signatures to an operation?}
\label{sec:specificity}
The next question is whether a score fitted on one operation also
identifies that operation. To test this, we fit each score on one
operation against organic Twitter and evaluate it in two ways (Appendix Figure~\ref{fig:rep-task}): against organic
accounts, which any operation's accounts should outrank, and against the
other operations, which only the fitting operation's accounts should
outrank.

\paragraph{Action--gap policies.}
Every policy ranks every other Twitter operation above organic accounts
(AUC 0.96--1.0) but ranks its own operation above the others poorly
(0.24--0.80). Under the Venezuelan and Iranian policies, IRA accounts score
higher than the accounts the policy was fitted on (0.24 and 0.29). A policy
can, in other words, separate organized from organic production without
identifying the organization, which is consistent with the shared,
differentiated work reported across state-linked operations
\citep{starbird2019,king2017}.

\begin{figure*}[t]
\centering
\begin{minipage}[t]{0.305\linewidth}
\centering\vspace{0pt}
\makebox[\linewidth][l]{\sffamily\fontsize{9}{11}\selectfont A}\par\smallskip
\includegraphics[height=1.8in, width=\linewidth, keepaspectratio]{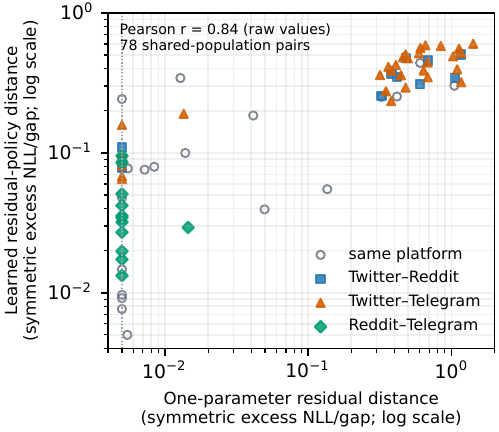}
\end{minipage}\hfill
\begin{minipage}[t]{0.365\linewidth}
\centering\vspace{0pt}
\makebox[\linewidth][l]{\sffamily\fontsize{9}{11}\selectfont B}\par\smallskip
\includegraphics[height=1.8in, width=\linewidth, keepaspectratio]{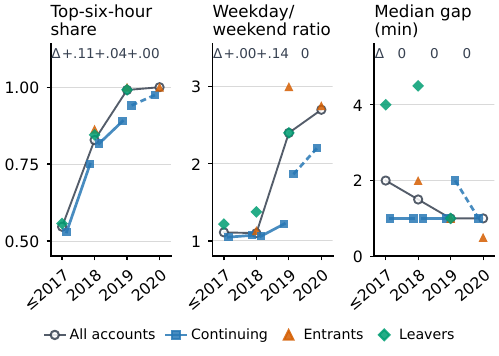}
\end{minipage}\hfill
\begin{minipage}[t]{0.29\linewidth}
\centering\vspace{0pt}
\makebox[\linewidth][l]{\sffamily\fontsize{9}{11}\selectfont C}\par\smallskip
\includegraphics[height=1.8in, width=\linewidth, keepaspectratio]{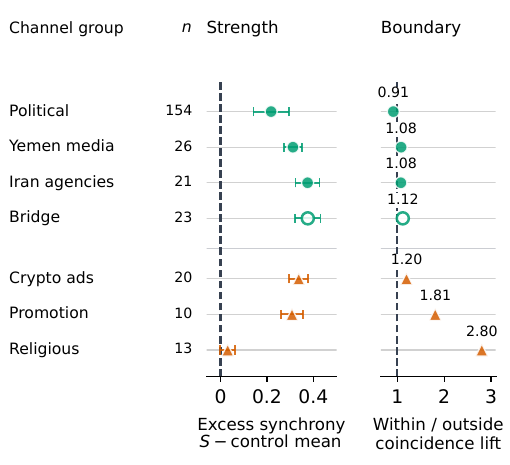}
\end{minipage}
\caption{(A) Learned gap-policy distance against one-rate tempo distance
for 78 pairs of 13 populations ($r=\publicTPPPearson$,
$\rho=\publicTPPSpearman$; excess NLL per held-out gap, log axes, floor
0.005); color and shape give the platform pair. (B) Top-six-hour share,
weekday/weekend ratio, and median gap (minutes) in the four-period Twitter archive, for all, continuing, entering, and leaving
accounts. (C) Telegram groups: original-post synchrony in ten-minute bins
above the volume-matched control mean (bars: s.d.\ across control sets).}
\label{fig:context-signatures}
\end{figure*}

\paragraph{Phase.}
Phase discrimination depends on the operation pair and the classifier.
The linear IRA score separates the IRA from Venezuela (0.86) and China
(0.66), is at chance against Iran (0.52), and ranks GRU accounts above IRA
accounts (0.32). Gradient boosting separates the IRA from Iran (0.85)
despite their overlapping working-hours windows, and from the other
operations at 0.81--0.91, while still ranking Iranian and GRU accounts
above organic accounts (0.83 and 0.91). Training on both operations'
labels separates all six pairs (0.91--0.97 linear, 0.95--1.0 gradient
boosting); for IRA--Iran the linear AUC rises from 0.52 to 0.91 (Appendix
Table~\ref{tab:rep-task-t3}). Profiles without phase (i.e., hourly
concentration, weekday structure, tempo, activity, and action mix) also
separate all six pairs (0.93--0.99). Descriptively, the IRA combines
broadcasting with weekday concentration, Venezuela has a later UTC window,
and Iran an inverted weekday/weekend ratio (Figure~\ref{fig:ops}A, B).
These differences may reflect institutional schedules
\citep{zerubavel1981}, but archive, period, role, and reference
differences remain possible explanations.

\paragraph{Tempo and action structure.}
Tempo does not transfer to Reddit.
Fitting a reference on each
platform, we find that Twitter operations post faster than their
reference, whereas the Reddit IRA and the Telegram keyword groups post at
roughly the pace of theirs (Figure~\ref{fig:ops}D). The relative-tempo
residual ranks the 64 eligible IRA accounts against volume-restricted
Reddit controls no better than chance (AUC 0.50), and even on Twitter the
tempo densities of the four operations only reach 0.33--0.59.
Tempo also explains what the learned policies pick up: across 78 pairs drawn from 13
populations, gap-policy distance closely follows a one-rate tempo
comparator ($r=\publicTPPPearson$, $\rho=\publicTPPSpearman$;
Figure~\ref{fig:context-signatures}A). But not all of it. The GRU archive
posts at an unremarkable rate, 3.6 posts a day with a 51-minute median
gap, and still scores 0.96--1.0 against organic accounts under every
policy (Appendix Figure~\ref{fig:rep-task}). Its burstiness (0.57, against
0.16 for organic accounts) and 86\% original-post share indicate that the
actions and gaps carry information beyond average
posting speed.

\paragraph{Same-region comparison.}
Does the timing gap survive once we control for how much accounts post
and what they post? We set Russian-language IRA accounts
against the Current Policy network, a separately disclosed Russian
operation \citep{twitter2020currentpolicy}, and match accounts within
each year on event count, active days, observation span, and quarterly
activity shares. We then compare hourly concentration, posting gaps, and
burstiness, leaving daily peak times aside (see the regional comparison
appendix). For 2016, 173 pairs survive matching. One difference is that IRA accounts have a lower median share of original
posts, 0.56 against 0.76. When we also match on action shares, 79 pairs
remain and the timing gap shrinks but does not close. The IRA-minus-Current
Policy difference in top-six-hour mass goes from $-6.6$ to $-5.3$
percentage points, and the median-gap difference drops from 62 to 18
minutes. Part of what looked like a timing difference, then, is a
difference in what accounts were doing, in line with the differentiated
production described by \citet{linvill2020troll} and \citet{starbird2019}. Some differences between the groups remain after matching, so we cannot
tell whether the observed gap reflects organizational routines or other
group differences.

%% file: sections/07_rq3_context.tex
\subsection{RQ3: How do signatures change within an operation?}
\label{sec:component-analysis}
To address RQ3, we examine whether a signature changes when the platform
is held fixed, using (i) four periods of one Twitter archive, (ii) IRA
cohorts on both platforms, and (iii) individual accounts linked across
platforms.

\paragraph{Schedule and pace over four periods.}
\label{sec:china}
The longest Twitter archive in Table~\ref{tab:data} spans four periods on
one platform, long enough to follow the same accounts from one period to
the next. Figure~\ref{fig:context-signatures}B separates each period's
population into accounts that continue, enter, and leave. Top-six-hour
mass rises from 0.55 through 2017 to 0.83 in 2018, 0.99 in 2019, and 1.00
in 2020, and continuing accounts move with it: among 209 accounts active
in both 2017 and 2019, the median paired change is $+0.106$, with 69\% of
accounts increasing. Entrants have more concentrated profiles than
continuing accounts, so replacement alone does not produce the rise. The
weekday/weekend ratio stays near 1.1 through 2018 and then reaches 2.4 and
2.7 (median paired change among continuing accounts $+0.14$ between 2018
and 2019). Pace changes less. The median paired change in accounts' median
gaps is zero in every adjacent contrast, while the 90th-percentile gap
lengthens for 63\% of accounts and burstiness falls for 72\% (Appendix
Table~\ref{tab:china-stayers}). Some components of production changed
while others did not, which is consistent with routines theory
\citep{feldman2003}, though staffing, tooling, instructions, and changed
activity levels remain possible explanations.

The two years do not contain equal event counts (median 593 events in
2017 and 399 in 2019, and only 37\% of accounts posted more in 2019), so
we checked whether sparser activity explains the tightening.
Re-estimating each account's hourly profile from the same number of
events in both years leaves the increase at $+0.099$, and a common
30-event budget for every account gives $+0.080$, so it does not, although
the increase is larger among accounts whose activity fell (Spearman
$-0.37$ with the change in log count). Note that the unchanged median gap
is one summary at one-minute resolution and not evidence of a stable pace:
individual accounts moved in both directions, the middle half by up to a
factor of three either way, and the tail changes track the fall in event
counts (Spearman $-0.63$ and $+0.36$). Only 34 of the 3,879 accounts
observed in 2019 remain eligible in 2020, so we read the final step at the
population level. In our own fits, gap-only policies separate the
through-2017 cohort from the 2019 and 2020 cohorts (AUC 0.83 and 0.89),
while the two later cohorts stay close to each other (0.66).

\paragraph{Audience cohorts.}
\label{sec:tempo}
The BLM and MAGA hashtag cohorts target opposite
audiences and have median inter-event gaps of 4.5 and one minute, compared
with 2,533 minutes in organic Twitter. Their phase alignment toward the
audience is unresolved: BLM's audience-minus-home cosine contrast is
$+0.106$ [$-0.007$, $0.207$] (Appendix Table~\ref{tab:adaptation}).
English-language IRA accounts have an hour profile closer to
Russian-language IRA than to organic Twitter (cosine 0.99 versus 0.85;
contrast $-0.135$ [$-0.153$, $-0.118$]). English- and Russian-language
accounts retain similar hours despite different audiences, consistent with a
distinction between audience-facing presentation and production traces
\citep{goffman1959}. 

\paragraph{Linked accounts.}
\label{sec:individuals}
Of 20 exact-handle Twitter--Reddit links, 11 are recoverable and six
eligible. Their mean profile distance is 3.486, versus 3.455 under 2,000
random relabelings of the same six targets ($p=0.549$); separately
archived ranks have median 63.5 among 96 candidates. One pair shifts
from 99\% original items on Twitter to 16\% on Reddit, illustrating that
population resemblance can coexist with changing account roles. This
small diagnostic provides no clear evidence of individual correspondence
and cannot establish account anonymity or common human ownership.

%% file: sections/08_telegram.tex
\section{Case Study: Production Signatures on Telegram}
\label{sec:telegram-main}
Two channels can follow the same daily schedule and never actually post
together. Telegram lets us pull these apart, since forward records show
which channels draw on the same sources (Figure~\ref{fig:pipeline}). We
cluster channels by the sources they forward from and ask whether their
original posts line up in time in ways that looking at each channel on
its own would not catch.
\begin{figure*}[t]
\centering
\includegraphics[width=\linewidth]{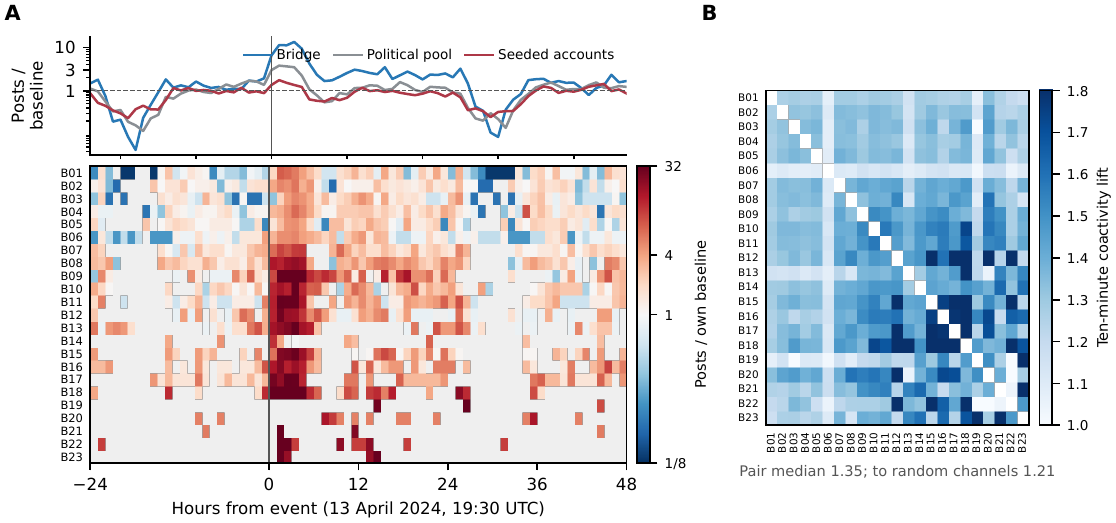}
\caption{Original-post activity of the 23 anonymized Telegram bridge
channels. (A) Hourly original posts around the 13 April 2024 Iranian
missile attack, each channel scaled by its mean hourly rate over the prior
four weeks; grey marks hours with no posts, the top strip pools each group
against its pooled baseline, and rows are ordered by baseline rate.
(B) How often each pair of channels posts in the same ten-minute bin over
47 weeks. Channels are labeled by rank.}
\label{fig:tg-traces}
\end{figure*}
\paragraph{Comparison groups.}
The unit of production on Telegram is the channel, a one-to-many feed
whose actions are original posts and forwards of other channels' posts,
the counterpart of reposts. A forward names its source, which the account
archives lack, so we use forwards to select comparison groups and the
original-post stream, with forwards excluded, to measure whether groups
post together. Channels are connected if they forward at least three of
the same external sources. Two compact components are associated with
Yemen conflict media ($n=26$) and Iranian agencies ($n=21$); the names
describe each component's most prominent affiliation, and membership is
mixed. Adding four matches to Iranian outlets linked by Iran-attributed
Twitter accounts to the Iranian-agency component gives a 23-channel
\emph{bridge}. Note that this cohort was selected retrospectively and that
its name describes an overlap in sources across platforms.

\paragraph{Individual signatures.}
The bridge's daily phase resembles Iran-attributed Twitter (cosine 0.965),
while its relative tempo is close to the Telegram reference (pooled
residual KS 0.031, against 0.42 for Iran--Twitter), the same pattern as
across Twitter and Reddit. The transferred phase + schedule model does not
isolate it either: at the logistic cutoff, 67.0\% of 91 seeded channels,
which carry the platform's verification badge, exceed it, against 43.5\%
of bridge channels, and bridge channels have shorter median gaps and less
weekday concentration than seeded channels (Telegram appendix). Without
labels this does not confirm the RQ1 result, but it repeats its shape on
a third platform. Individual signatures cannot show whether these channels
act together, which the forwarding records let us test.

\paragraph{Synchronized-posting statistic.}
For group $G$, let $N^{(b)}_{ut}$ count channel $u$'s original posts in time
bin $t$ of width $b$. Excluding forwards from the outcome, we define
\begin{equation*}
S_G(b)=
\frac{\operatorname{Var}_t \sum_{u\in G} N^{(b)}_{ut}}
{\mathbb{E}_{\pi}\left[\operatorname{Var}_t \sum_{u\in G}
N^{(b)}_{u,t+\pi_u}\right]},
\end{equation*}
where $\pi_u$ is a random whole-week circular offset in bins. The null
keeps each channel's weekly rhythm while shifting its activity relative to
the others, so $S=1$ means the aggregate variance matches its expectation
under these shifts, and values above one mean the channels post together
more than their weekly rhythms imply. Note that the null does not account
for shared responses to news events.

\paragraph{Synchronized posting in shared-source groups.}
Shared-source groups post together well beyond their weekly rhythms
(Figure~\ref{fig:context-signatures}C). The bridge's ten-minute score is
$S=1.482$, against $1.106\pm0.054$ across 100 volume-matched political
control sets; the Yemen-associated and Iranian-agency components reach
1.393 and 1.469 against control means of 1.081 and 1.094, 8.0 and 7.2
standard deviations above them. Political neighborhoods are not special
here: a crypto-advertising cluster ($n=20$) and a promotion-exchange
cluster ($n=10$) reach 1.483 and 1.371, while a religious-content cluster
($n=13$) stays near its reference, so posting together is no more specific
to political groups than the action--gap policies in RQ2 were to one
operation. Figure~\ref{fig:tg-traces} shows the bridge itself. In the
hours after the 13 April missile attack (A), its pooled original posting
reaches 13.5 times baseline, against 3.8 for the political pool and 1.8
for seeded accounts, across channels with high and low baseline rates.
Over 47 weeks (B), the median pair of bridge channels posts in the same
ten-minute bin 1.35 times as often as chance, against 1.21 for bridge
channels paired with random eligible channels. Forwards are excluded from
both readouts, and only 0.8\% of bridge members' forwards come from
another bridge channel (Appendix, ``Forwarding among bridge members''), so
the channels publish together far more than they republish one another.
In Goffman's terms, a forward is a declared tie and posting in the same
window is what production gives off \citep{goffman1959}; it may trace a
shared desk, a shared feed, or a shared news environment.

\paragraph{Synchronized posting by timescale.}
Comparing ten-minute and one-minute resolution narrows the
interpretation. Across all 1,162 eligible channels, $S$ is 1.052 at ten
minutes and 1.041 at one minute, and the seeded accounts sit at 1.019 and
1.007, so the platform as a whole shows little excess at either scale.
The bridge and the Iranian-agency component keep their excess mainly at
the coarser scale (1.482 and 1.469 at ten minutes against 1.100 and 1.097
at one minute, where the bridge still exceeds its matched control mean of
1.035 by a small margin). The Yemen-associated component retains more at
one minute (1.393 against 1.230), so its channels come closer to posting
in the same minute, and the fast-tempo tail of the platform shows none
(1.004). The shared-source groups publish within the same few minutes
rather than at the same instant, suggesting channels responding to a
common feed rather than automated simultaneous posting.

\paragraph{Group boundaries.}
Several arrangements could produce this pattern. Staffed shifts,
scheduling tools, queues, and role allocation are the obvious candidates
\citep{starbird2019,keller2020,king2017,schoch2022}, but so is ordinary
newsroom practice: outlets watch one another and publish the same stories
within minutes without sharing any staff \citep{boczkowski2010}, and a
shared news environment gives everyone in it a common clock
\citep{zerubavel1981}. The boundary comparison in
Figure~\ref{fig:context-signatures}C helps tell these apart. In the Yemen-associated and Iranian-agency
components, channels post together with matched outsiders almost as
often as with each other (within-to-outside ratios of about 1.08), which
is what a common clock would look like. The crypto-advertising and
promotion-exchange clusters show the same level of synchronized posting,
but it stays inside the cluster (1.20 and 1.81). The
statistic separates coordination that an environment imposes from
coordination that belongs to a group, even if neither comparison names
the organization or the mechanism behind it. The case study adds a relational component to production
signatures: how closely synchronized posting follows group boundaries.

%% file: sections/09_conclusion.tex
\section{Conclusion}
\label{sec:conclusion}

We modeled timing and action traces as production signatures and asked
which components carry across platforms, whether they distinguish
operations, and how stable they are over time. A model trained only on
Twitter ranks IRA accounts on Reddit at AUC 0.90--0.94 with no Reddit
labels. Daily phase carries most of this transfer; volume reverses between
platforms and action mix follows platform role. The signatures separate
the operations we studied from organic activity more reliably than they
identify which operation produced an account. Over four periods, one
operation's schedules became more concentrated while its median pace did
not change, and on Telegram a group synchrony statistic detected channels
publishing together that individual timing missed. Production signatures
transfer as screening signals but do not establish who runs an account. In future work we plan to pair production signatures with content
and network signals, test whether they survive deliberate changes in
timing, and evaluate them on more operations across different
platforms.

%% file: sections/appendix.tex
\newpage
\section{Appendix}

\subsection{Limitations and Ethical Considerations}
\label{sec:limitations}
\label{app:extended-limitations}

\paragraph{Limitations.}
Our results should be considered in light of several limitations. First,
the labeled cross-platform evidence concerns one operation, the IRA, on
Twitter and Reddit; Iran, Venezuela, China, and GRU test generalization
across operations within Twitter, not additional platform-transfer cases.
Second, recrawl survivorship favors higher-volume accounts, so the Reddit
estimand is the recoverable subset of the named accounts (944 listed, 146
recovered, 96 eligible), and all comparisons concern detected and removed
campaigns. Third, volume matching is coarse and Twitter controls come from a
predominantly Western, English-heavy crawl, so residual volume, role, and
geography differences remain. Fourth, Telegram groups are co-forward
neighborhoods, not platform-labeled operations, so transferred-score
exceedance rates do not estimate Telegram false positives. Finally, the
within-operation and linked-persona analyses rest on small samples (34
accounts in the final China period, six exact-handle pairs).

\paragraph{Ethical Considerations.}
Content-agnostic signals support the integrity of online discourse across
platforms, but the schedules our models read are shared by ordinary
institutions, so misapplication could label organic accounts that keep
regular hours as inauthentic. Our models use no content and do not identify individuals, all
operation data come from public platform disclosures, Telegram channels
are anonymized, and signature similarity alone does not establish a common
operator.

\subsection{Representation-by-Task Protocol}
\label{app:rep-task}
\begin{figure*}[!t]
\centering
\includegraphics[width=\linewidth]{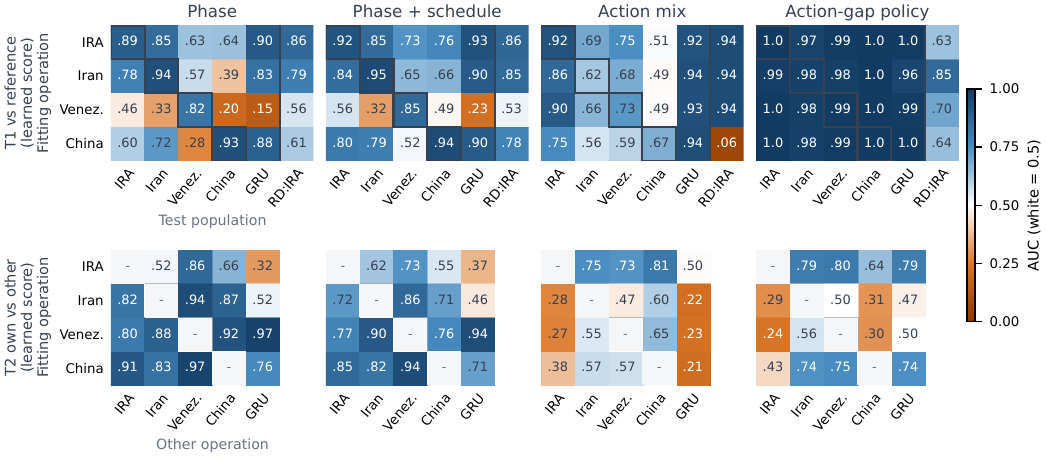}
\caption{Operation-by-test-population score matrices for four
representations (logistic regression for profiles; log-likelihood ratio
for the action--gap policy). Rows: the operation whose training accounts
fitted the learned score. T1: AUC against the test population's platform
reference (organic Twitter; 168 volume-matched organic Reddit accounts for
RD:IRA); outlined cells are the fitting operation's held-out accounts. T2:
own-versus-other AUC with the fitting operation as the positive class.
Blue is above 0.5, orange below, white 0.5; dashes mark self-comparisons.}
\label{fig:rep-task}
\end{figure*}

Figures~\ref{fig:transfer-gap} and~\ref{fig:rep-task} and
Table~\ref{tab:rep-task-t3} use one 70/30 account split with a fixed seed
(IRA 2,042/875, Iran 344/148, Venezuela 521/223, China 2,405/1,030,
organic Twitter 1,220/523); GRU (30 accounts) and the Reddit cohort (96
IRA, 643 organic, 168 volume-matched) are test-only. T1 applies each
learned score to a test population against its platform reference; T2 ranks
the fitting operation against each other operation; T3 refits on the two
operations' training accounts or, for the policy, uses the ratio of the two
fitted policies. Every cell carries a 500-draw account-bootstrap interval.
The same split and cohorts serve the Table~\ref{tab:main-results} baselines
that we fit ourselves: BLOC and the reimplementation of
\citet{alizadeh2020} are fitted on the Twitter training accounts, frozen,
and scored on the 96 Reddit IRA accounts against the 168 matched controls,
while the PS rows use 161 matched controls from a separate draw. Within
Reddit, the policy row uses five-fold cross-validation on the 96/643
cohort with the log-likelihood ratio standardized per fold on the training
organic accounts; the other within-Reddit rows use ten repeats of the same
shared five-fold splits. Fitted on Iran or China instead of the IRA, BLOC
reaches 0.909 and 0.804 on the same matched Reddit cohort.

\begin{table}[!t]
\centering\footnotesize
\setlength{\tabcolsep}{0pt}
\input{macros/rep_task_t3_table.tex}
\caption{Supervised pairwise operation classification (T3): held-out AUC
of a scorer fitted on both operations' training accounts. LLR, ratio of the
two fitted policies.}
\label{tab:rep-task-t3}
\end{table}

\subsection{Transfer Evaluations and Component Ablation}
\label{app:ablation}

\begin{table}[!t]
\centering\footnotesize
\setlength{\tabcolsep}{0pt}
\begin{tabular*}{\columnwidth}{@{\extracolsep{\fill}}lccc@{}}
\toprule
Route & \(n_+/n_-\) & \shortstack{Full profile\\(LR / GBM)} &
\shortstack{Phase + weekday\\(LR / GBM)} \\
\midrule
TW14 $\rightarrow$ RD-pol. & 69/491 & 0.79 / 0.74 & 0.90 / 0.95 \\
TW14 $\rightarrow$ RD-crypto & 25/467 & 0.93 / 0.89 & 0.80 / 0.90 \\
TW16 $\rightarrow$ RD-pol. & 69/491 & 0.85 / 0.76 & 0.85 / 0.92 \\
TW16 $\rightarrow$ RD-crypto & 25/467 & 0.94 / 0.88 & 0.79 / 0.91 \\
Full TW $\rightarrow$ RD & 96/663 & --- / 0.87 & 0.88 / --- \\
RD $\rightarrow$ TW & 2801/443 & --- / 0.74--0.80 & 0.90 / --- \\
\midrule
Volume only & 96/663 & \multicolumn{2}{c}{0.125 (reversed ordering)} \\
\bottomrule
\end{tabular*}
\caption{Cross-platform operation-versus-organic AUC (logistic regression /
gradient boosting) with no target-platform labels. TW14 and TW16 are the
2014 and 2016 Twitter source windows; ``pol.'' is the Reddit political
cohort. These routes use the earlier phase + weekday specification and an
unmatched 663-account organic pool; the common-cohort, volume-matched
results that Table~\ref{tab:main-results} reports are in the component
ablation below. After volume matching, phase + weekday AUC remains 0.83 and
0.91.}
\label{tab:transfer}
\end{table}

\begin{figure*}[!t]
\centering
\includegraphics[width=.90\linewidth]{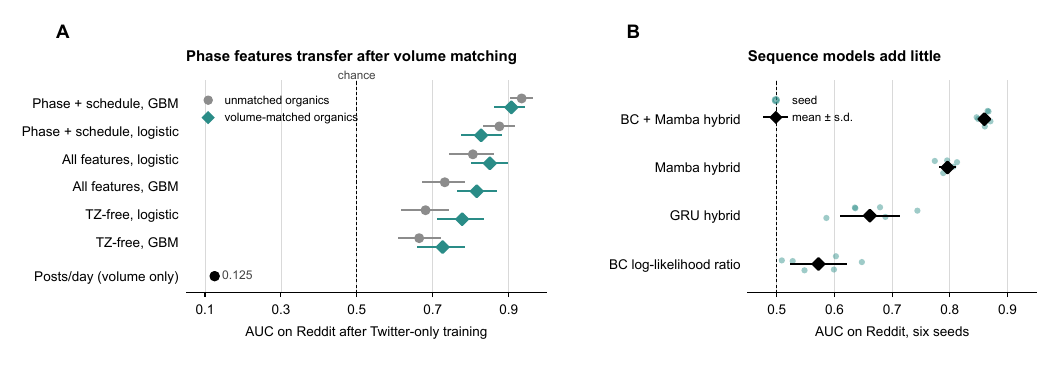}
\caption{(A) Twitter-trained feature-model AUC on Reddit against unmatched
(grey) and volume-matched (teal) controls, with 95\% bootstrap CIs; the
posts-per-day baseline reverses. (B) Six-seed results for sequence and
hybrid models (mean $\pm$ s.d.).}
\label{fig:detectors}
\end{figure*}

\paragraph{Component ablation.}
On the gap-eligible cohort (at least 30 IRA / 50 organic distinct
timestamps; 64/599 accounts, 124 matched controls), tempo rises to
0.85/0.83 and phase + schedule to 0.93/0.93 (LR/GBM), so the advantage of
phase + schedule over tempo persists but narrows. Matched gradient-boosting AUC is $0.912\pm0.012$ at 5\% and
$0.917\pm0.010$ at 10\% of the source accounts. Flagged organic accounts under the linear scorer have median weekday
ratio 4.0 versus 1.1 and median gap 26 versus 71 minutes; missed IRA
accounts are the least schedule-concentrated in the cohort. Figure~\ref{fig:detectors}B reports the six-seed sequence and hybrid
models, none of which exceeds the learned profile transfer.

\paragraph{Reference sensitivity of relative tempo.}
\label{sec:tpp}
Twitter operations occupy a short-residual regime (means 0.31--0.45) and
separate from held-out organic accounts at AUC 0.933--0.990, whereas the
six held-out Reddit IRA accounts have mean residual 0.997, near the Reddit
reference. Over all 64 eligible Reddit IRA accounts, mean relative-tempo AUC is
0.26 against the original heavy-commenter reference and 0.50 after
restricting organics to 30--1,000 actions.

\subsection{Regional Comparison: IRA and Current Policy}
\label{sec:regional-signature}
We compare the June 2020 Current Policy archive
\citep{twitter2020currentpolicy} with the IRA release in calendar 2016
(2015 and 2017 as sensitivity windows), requiring at least 100 events, 14
active UTC days, and a Russian-language label on at least 60\% of events.
Accounts are paired without replacement within 0.25 $\log_{10}$ units of
event count, active days, and span, and within 0.25 of each quarterly
share; the role variant also caps action-share differences at 0.20.
Residual imbalance remains (standardized mean differences up to 0.35), a
stricter balance check retains only 27 and 13 pairs, and the 2017
contrasts are null.

\subsection{China Period Comparison}
\label{sec:china-detail}

\begin{table}[!htbp]
\centering\small
\setlength{\tabcolsep}{2.5pt}
\begin{tabular}{@{}lrlc@{}}
\toprule
Period comparison & Stayers & $\Delta$hour mass [95\% CI] & $\Delta$gap \\
\midrule
$\leq$2017--2018 & 379 & +.114 [.082, .131] & 0 \\
2018--2019 & 673 & +.037 [.028, .048] & 0 \\
$\leq$2017--2019 & 429 & +.261 [.227, .298] & $-1$ [$-2$, $-1$] \\
2017--2019 & 209 & +.106 [.078, .154] & 0 [0, 1] \\
\bottomrule
\end{tabular}
\caption{Paired changes within continuing China accounts: medians of
top-six-hour mass and median gap (minutes) with 2,000-draw
account-bootstrap CIs. The last row compares equal twelve-month windows.}
\label{tab:china-stayers}
\end{table}

Eras are defined by posting date (through 2017, 2018, 2019, 2020 onward);
an account contributes to an era if it has at least 30 events there. For
the 209 accounts with both calendar-2017 and calendar-2019 cells,
re-estimating the hourly profile at equal event counts leaves the paired
increase at $+0.099$ [0.071, 0.147], and a common budget of 30 or 100
events gives $+0.080$ [0.047, 0.109] and $+0.082$ [0.047, 0.116]. The
2019$\rightarrow$2020 step has 34 stayers against 3,845 leavers and 643
entrants.

\subsection{Audience Cohorts}
\begin{table}[!htbp]
\centering\small
\setlength{\tabcolsep}{3.5pt}
\begin{tabular}{@{}lrrrr@{}}
\toprule
Cohort & Accounts & Gap (min) & Org. cosine & ru-IRA cosine \\
\midrule
\publicAdaptationRows
\bottomrule
\end{tabular}
\caption{IRA cohort phase and tempo. Cosines compare equal-account-weight
hour centroids; gaps are medians of account medians; Org.\ is the
1,743-account organic Twitter reference.}
\label{tab:adaptation}
\end{table}
Language cohorts require at least 60\% of tweets in the corresponding
language; topic cohorts require at least 50 archive tweets with at least
10\% using the BLM or MAGA hashtag lists. A 2,000-draw account bootstrap
gives BLM's audience-minus-home cosine contrast $+0.106$ [$-0.007$,
$0.207$], MAGA's $-0.002$ [$-0.074$, $0.079$], and English-IRA's $-0.135$
[$-0.153$, $-0.118$].

\subsection{Telegram: Construction, Controls, and Forwarding Network}
\label{sec:telegram-sync}

\begin{figure*}[!t]
\centering
\includegraphics[width=.92\linewidth]{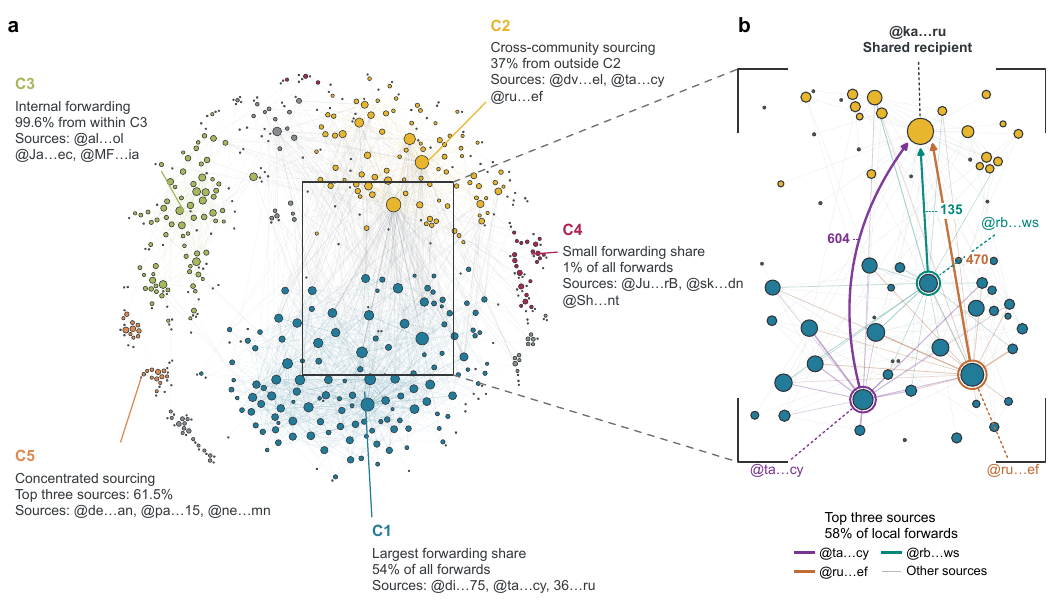}
\caption{Telegram forwarding network. (a) Largest weakly connected
component (578 channels, 2,297 forwarding pairs). Node size is degree,
colors mark the five largest Louvain communities (C1--C5), and edge width
scales with log forward volume; lists rank each community's top sources by
forwards received. (b) The boxed region enlarged: colored rings mark its
three leading sources and arrows their strongest source-to-recipient flows,
with forward counts. Handles are partially masked.}
\label{fig:tg-forward-graph}
\end{figure*}

\paragraph{Bridge construction.}
Channels are connected if they forward at least three of the same external
sources. Iran-attributed Twitter accounts link 836 times to six Iranian
outlets; four name matches on Telegram, combined with the 21-channel
Iranian-agency co-forward component (two matches lie inside it), give the
$4+21-2=23$ bridge channels. The later-period check ($S=1.385$ against
$1.101\pm0.066$) reuses this cohort and so assesses temporal sensitivity
only.

\paragraph{Controls and estimates by timescale.}
Each group is compared with 100--200 volume-matched control sets,
political-only where the pool allows; control summaries are means $\pm$ s.d.\ across sampled sets. Against
political-only controls, the Yemen-associated component ($1.393$) is 8.0
standard deviations above the control mean ($1.081\pm0.039$) and the
Iranian-agency component ($1.469$) 7.2 above ($1.094\pm0.052$). The bridge's
one-minute estimate ($1.100$) remains above its matched control mean
($1.035\pm0.022$).

\paragraph{Forwarding among bridge members.}
Figure~\ref{fig:tg-forward-graph} shows Louvain communities (resolution 1,
seed 11) on the shard's forwarding adjacency: 54\% of the component's
49,895 forwards are received by C1. Of the bridge's 27,315 forwards in
2024, 224 (0.8\%) originate from another bridge channel; the rest come from
507 outside origins, 397 of which feed a single member.

\paragraph{Transferred score cutoffs.}
The phase + schedule comparison uses 2024 original posts (at least 20 per
channel) and the organic Reddit 95th-percentile cutoffs (0.9702 logistic,
0.9714 gradient boosting): 67.0\% of 91 seeded channels exceed the logistic
cutoff against 43.5\% of the 23 bridge channels (gradient boosting 39.6\%
and 21.7\%). Bridge channels have median gap 8.6 versus 24 minutes.

\subsection{Feature Definitions and Reproducibility}
\label{app:repro}

\paragraph{Feature definitions.}
Phase: 24 UTC-hour proportions and the masses of three fixed windows,
06--15 UTC (Moscow working hours), 00--05 UTC (night), and 15--20 UTC
(evening). Schedule: the ratio of mean weekday to mean weekend daily counts,
capped at 5; the summed proportion of the six largest hours; hour entropy;
and active days divided by span days. Tempo: median and 90th-percentile gap
in minutes, burstiness $(\sigma-\mu)/(\sigma+\mu)$ of gaps, and the share of
gaps under one minute. Action mix: the three action proportions. Volume,
used only as a separate baseline, is posts per active day. Gap and volume
features enter as $\log(1+x)$; all features are standardized on the source
training accounts. Numerical safeguards for the interval-adaptive model are
the constants stated in the Methodology: a $10^{-3}$-minute floor on the
left interval endpoint, the clipping of $U_{uj}$ to $[10^{-9},1-10^{-7}]$,
and a $10^{-8}$ floor on residuals.

\paragraph{Models and compute.}
Profile classifiers are $\ell_2$ logistic regression (balanced class
weights, up to 2,000 iterations) and scikit-learn default gradient boosting
(100 estimators, depth 3) on source-standardized features. Policies and the
conditional gap model are two-layer Mamba encoders (width 128 and 64; the gap
model with a $K=8$ log-normal mixture head and 32 log bins), trained with
AdamW at learning rate $10^{-3}$, weight decay 0.01, batch 48, 512-token
windows, three epochs (a 150-step minimum for the small within-Reddit
policy fits). Content baselines are mean-pooled multilingual MPNet embeddings
with logistic regression, a fine-tuned XLM-R base model (one epoch, learning
rate $2\times10^{-5}$, account-disjoint split), and the \citet{alizadeh2020}
reimplementation (81 per-post features in their five groups, 1,000-tree
random forest, up to 100 posts per account). BLOC uses character 1--2-gram
TF-IDF over action and pause symbols with balanced logistic regression. All
experiments ran on one server (36 CPU cores, one NVIDIA A40 GPU at a time)
in under 40 GPU-hours. 

%% file: macros/rep_task_t3_table.tex
\begin{tabular*}{\columnwidth}{@{\extracolsep{\fill}}lcccccc@{}}
\toprule
Representation & \shortstack{IRA\\Iran} & \shortstack{IRA\\Venez.} & \shortstack{IRA\\China} & \shortstack{Iran\\Venez.} & \shortstack{Iran\\China} & \shortstack{Venez.\\China} \\
\midrule
Phase, LR & 0.91 & 0.92 & 0.96 & 0.96 & 0.97 & 0.97 \\
Phase, GBM & 0.95 & 0.97 & 0.98 & 1.00 & 0.98 & 0.99 \\
Phase + sched., LR & 0.91 & 0.94 & 0.96 & 0.97 & 0.97 & 0.97 \\
Phase + sched., GBM & 0.96 & 0.98 & 0.99 & 1.00 & 0.99 & 0.99 \\
Action mix, LR & 0.75 & 0.73 & 0.82 & 0.54 & 0.69 & 0.69 \\
Action mix, GBM & 0.81 & 0.81 & 0.88 & 0.72 & 0.75 & 0.77 \\
Policy, LLR & 0.84 & 0.98 & 0.98 & 0.80 & 0.87 & 0.97 \\
\bottomrule
\end{tabular*}